\documentclass[runningheads]{llncs}

\usepackage{eccv}

\usepackage{eccvabbrv}

\usepackage{graphicx}
\usepackage{booktabs}
\usepackage{multirow}
\usepackage[ruled,vlined]{algorithm2e}
\usepackage{subcaption}

\usepackage{xcolor}
\usepackage{circuitikz}
\usepackage{tikz}
\usepackage{braket}

\usepackage[accsupp]{axessibility}  % Improves PDF readability for those with disabilities.

\usepackage[pagebackref,breaklinks,colorlinks,citecolor=eccvblue]{hyperref}
\usepackage{orcidlink}

\begin{document}

\title{WTHaar-Net: a Hybrid Quantum-Classical Approach}

\titlerunning{WTHaar-Net: Hybrid Quantum-Classical Approach}

% Include the authors' ORCID for the camera-ready version, if at all possible.
\author{
Vittorio Palladino\inst{1,2}$^{\star}$ \and
Tsai Idden\inst{1}$^{\star}$ \and
Ahmet Enis Cetin\inst{1}$^{\dagger}$
}

\authorrunning{V.~Palladino et al.}

\institute{
University of Illinois Chicago (UIC), Chicago, IL, USA
\and
Politecnico di Milano, Department of Electronics, Information and Bioengineering, Milano, Italy\\
\email{vpall3@uic.edu, itsai@uic.edu, aecyy@uic.edu, vittorio.palladino@mail.polimi.it}
}

\maketitle

\footnotetext{
$\star$ Equal contribution,
$\dagger$ Corresponding author,
}

\begin{abstract}
Convolutional neural networks rely on linear filtering operations that can be reformulated efficiently in suitable transform domains. At the same time, advances in quantum computing have shown that certain structured linear transforms can be implemented with shallow quantum circuits, opening the door to hybrid quantum–classical approaches for enhancing deep learning models.

In this work, we introduce \textbf{WTHaar-Net}, a convolutional neural network that replaces the Hadamard Transform used in prior hybrid architectures with the Haar Wavelet Transform (HWT). Unlike the Hadamard Transform, the Haar transform provides spatially localized, multi-resolution representations that align more closely with the inductive biases of vision tasks. We show that the HWT admits a quantum realization using structured Hadamard gates, enabling its decomposition into unitary operations suitable for quantum circuits.

Experiments on CIFAR-10 and Tiny-ImageNet demonstrate that WTHaar-Net achieves substantial parameter reduction while maintaining competitive accuracy. On Tiny-ImageNet, our approach outperforms both ResNet and Hadamard-based baselines. We validate the quantum implementation on IBM Quantum cloud hardware, demonstrating compatibility with near-term quantum devices.
\end{abstract}


\section{Introduction}
\label{sec:intro}

Recent advances in quantum computing hardware have made it increasingly feasible to execute selected components of modern machine learning pipelines on near-term quantum processors. However, fully convolutional layers requires thousands of qubits that scales with the input dimensionality, quickly exceeding the capabilities of current devices. As a result, hybrid quantum-classical models have been used in prior work as a practical approach for exploiting quantum acceleration while retaining the scalability of classical deep learning.

A key insight from previous work on Hadamard Transform (HT)-based hybrid layers \cite{pan2023hadamardht, pan2021fast, pan2022block, pan2022ember, deveci2018energy}, is that certain structured linear transforms can be implemented efficiently on quantum hardware. In particular, the Hadamard Transform admits a simple circuit realization using only Hadamard gates and enables convolution-like computation through a transform-domain formulation \cite{agaian2002quantum, marcandelli2025partitioned, grigoryan2019paired}.

In this work, we build on the same principle but move beyond global Hadamard-domain mixing to a wavelet-domain representation. Specifically, we introduce \textbf{WTHaar-Net}, a hybrid architecture that replaces the global mixing induced by the Hadamard Transform with the \textbf{Haar Wavelet Transform}. Unlike the Hadamard Transform, which mixes all input components uniformly, the Haar transform provides a multi-resolution and spatially localized representation that better matches the inductive biases of vision tasks.
\begin{subequations}\label{eq:haar_hadamard}

\paragraph{Haar transform.}
The Haar transform is defined recursively as
\begin{align}
\mathrm{Haar}_{k+1}
&=
\begin{pmatrix}
\mathrm{Haar}_k \otimes \begin{pmatrix} 1 & 1 \end{pmatrix} \\[6pt]
2^{k/2} I_{2^k} \otimes \begin{pmatrix} 1 & -1 \end{pmatrix}
\end{pmatrix},
\end{align}
where
\begin{itemize}
\item $\mathrm{Haar}_k \in \mathbb{R}^{2^k \times 2^k}$ is the Haar transform matrix at level $k$,
\item $I_{2^k}$ is the $2^k \times 2^k$ identity matrix,
\item $\otimes$ denotes the Kronecker (tensor) product,
\item $2^{k/2}$ is a normalization factor.
\end{itemize}
The base case is
\begin{align}
\mathrm{Haar}_1
&=
\begin{pmatrix}
1 & 1 \\
1 & -1
\end{pmatrix}.
\end{align}

\paragraph{Hadamard transform.}
The Walsh--Hadamard transform satisfies the recursion
\begin{align}
\mathrm{Hadamard}_{2^{k+1}}
&=
\begin{pmatrix}
H_{2^k} & H_{2^k} \\
H_{2^k} & -H_{2^k}
\end{pmatrix},
\end{align}
where
\begin{itemize}
\item $\mathrm{Hadamard}_{2^k} \in \mathbb{R}^{2^k \times 2^k}$ is the Hadamard matrix of order $2^k$.
\end{itemize}
The base case is
\begin{align}
\mathrm{Hadamard}_2
&=
\begin{pmatrix}
1 & 1 \\
1 & -1
\end{pmatrix}.
\end{align}

\paragraph{Hadamard gate.}
The single-qubit Hadamard gate is
\begin{align}
H
&=
\begin{pmatrix}
1 & 1 \\
1 & -1
\end{pmatrix},
\end{align}
where
\begin{itemize}
\item $H$ acts on a single qubit,
\item its matrix representation coincides with the $2 \times 2$ Hadamard and Haar transform.
\end{itemize}

\end{subequations}

We empirically demonstrate that adopting the Haar wavelet leads to a better accuracy-efficiency trade-off. Across experiments on CIFAR-10 and Tiny-ImageNet, WTHaar-Net achieves a significant reduction in multiply-accumulate operations (MACs) compared to standard CNN baselines, while matching or improving classification accuracy.

This efficiency is enabled by the structure of the Haar Wavelet Transform (HWT), which is composed of simple, hardware-friendly operations such as pairwise sums and differences followed by structured permutations. We specifically choose the Haar wavelet due to the orthogonality of its transform matrix, which makes it well suited for both classical and quantum implementations. Notably, in the two-dimensional case, the $2 \times 2$ Haar matrix is identical to the Hadamard matrix, meaning that both transforms could be reproduce with the application of Hadamard gates.
This structure enables the efficient implementation of the Haar wavelet transform using Hadamard gates, by decomposing the recursive Haar matrix into a sequence of simpler unitary operations. For an input of length $N = 2^{m}$, the classical fast Haar wavelet transform runs in $\mathcal{O}(N)$ time.

Our main contributions are summarized as follows:
\begin{itemize}
\item \textbf{HWT-based hybrid pipeline:} We integrate the Haar Wavelet Transform as a front-end transform within a hybrid quantum-classical convolutional neural network.
\item \textbf{Quantum-friendly realization:} We present a quantum decomposition of the HWT using structured Hadamard gates, yielding circuits compatible with near-term quantum hardware constraints.
\item \textbf{Efficiency and accuracy:} Across CIFAR-10 and Tiny-ImageNet, WTHaar-Net achieves up to 44\% MAC reduction relative to standard CNNs while maintaining or improving accuracy, and consistently outperforms baselines.
\item \textbf{Hardware validation:} We demonstrate practical feasibility on real quantum hardware through patch-wise experiments on MNIST, implemented and evaluated on IBM Quantum cloud devices.

\end{itemize}
The implementation will be made available in a public repository following publication.
\section{Related Work}
\label{sec:related}

\paragraph{Wavelet-based CNNs.}
Wavelets provide multi-resolution representations with joint spatial and frequency localization, making them well suited for vision tasks. Prior work has integrated discrete wavelet transforms into CNNs as trainable components \cite{fujieda2018waveletcnn} or as downsampling operators in U-Net-style architectures \cite{liu2018mwcnn}. Yoo et al.\ \cite{yoo2021ldw} introduce learnable discrete wavelet pooling (LDW) that adaptively learns wavelet coefficients during training, achieving improved accuracy on CIFAR-10 while maintaining parameter efficiency. However, these methods operate purely in the classical domain and have not addressed quantum implementations. Our work bridges this gap by demonstrating that Haar wavelet transforms can be efficiently realized on quantum hardware while achieving competitive performance with classical wavelet-based approaches.

\paragraph{Hybrid quantum-classical neural networks.}
Hybrid quantum-classical models combine parameterized quantum circuits with classical learning pipelines to exploit near-term quantum devices \cite{benedetti2019parameterized,cerezo2021variational}. Quantum feature maps and variational circuits have been applied to classification tasks \cite{havlicek2019supervised,schuld2021machine, cong2019qcnn, liu2021hybrid}, and recent work has explored hybrid approaches combining quantum models with classical transforms such as PCA and Haar wavelets \cite{islam2023hybrid, marcandelli2025partitioned}\cite{cong2019qcnn}\cite{liu2021hybrid}. In vision, Hadamard Transform-based quantum convolutions offer computational efficiency but rely on global mixing, which limits spatial locality.

\paragraph{Quantum-friendly structured transforms.}
Real, orthogonal transforms with simple gate decompositions are attractive for hybrid quantum-classical pipelines. While quantum Fourier transforms are well studied \cite{nielsen2010quantum, weinstein2001qft, marcandelli2025partitioned, agaian2002quantum}, simpler structured transforms such as the Hadamard gate are more suitable for shallow circuits \cite{nielsen2010quantum, weinstein2001qft}. Motivated by this, we adopt the Haar wavelet transform, whose recursive sum–difference structure aligns with vision inductive biases and admits efficient quantum realizations. Our approach demonstrates how Haar-based feature extraction can be mapped to quantum circuits, bridging wavelet CNNs and hybrid quantum-classical vision models. \cite{shukla2023hybrid, liu2021hybrid}
\section{Preliminary}
We briefly review the fundamentals of the Haar wavelet transform that underpin the proposed approach.

\paragraph{\textbf{1D Haar Wavelet.}}
Let $\mathbf{x} \in \mathbb{R}^{N}$ be a one-dimensional signal of length $N = 2^{m}$:
\begin{equation}
\mathbf{x} =
\begin{bmatrix}
x_0 & x_1 & x_2 & x_3 & \cdots & x_{N-1}
\end{bmatrix}^{\top}.
\end{equation}

A single-level discrete Haar wavelet transform computes pairwise averages and differences:
\begin{equation}
a_k = \frac{1}{\sqrt{2}} \left( x_{2k} + x_{2k+1} \right), \qquad
d_k = \frac{1}{\sqrt{2}} \left( x_{2k} - x_{2k+1} \right),
\quad k = 0, \ldots, \frac{N}{2}-1 .
\end{equation}

The transformed vector is then given by
\begin{equation}
\mathbf{X} =
\begin{bmatrix}
a_0 & a_1 & \cdots & a_{\frac{N}{2}-1} &
d_0 & d_1 & \cdots & d_{\frac{N}{2}-1}
\end{bmatrix}^{\top}.
\end{equation}

This procedure is recursively applied to the approximation coefficients $\{a_k\}$, yielding a multilevel (multi-resolution) decomposition of the original signal {\bf x}.

\paragraph{Example.}
Consider a one-dimensional signal of length $N=4$,
\begin{equation}
\mathbf{x} =
\begin{bmatrix}
x_0 & x_1 & x_2 & x_3
\end{bmatrix}^{\top}.
\end{equation}
The first-level Haar transform produces
\begin{equation}
a_0 = \frac{1}{\sqrt{2}}(x_0 + x_1), \qquad
a_1 = \frac{1}{\sqrt{2}}(x_2 + x_3),
\end{equation}
\begin{equation}
d_0 = \frac{1}{\sqrt{2}}(x_0 - x_1), \qquad
d_1 = \frac{1}{\sqrt{2}}(x_2 - x_3).
\end{equation}

The transformed vector after one level is
\begin{equation}
\mathbf{X} =
\begin{bmatrix}
a_0 & a_1 & d_0 & d_1
\end{bmatrix}^{\top}.
\end{equation}

Applying the transform recursively to the approximation coefficients $(a_0, a_1)$ yields the second level:
\begin{equation}
a'_0 = \frac{1}{\sqrt{2}}(x_0 + x_1 + x_2 + x_3), \qquad
d'_0 = \frac{1}{\sqrt{2}}(x_0 + x_1 - x_2 - x_3).
\end{equation}

The final Haar wavelet representation is therefore
\begin{equation}
\mathbf{X}_{\mathrm{HWT}} =
\begin{bmatrix}
a'_0 & d'_0 & d_0 & d_1
\end{bmatrix}^{\top}.
\end{equation}

\begin{figure}[t]
    \centering
    \includegraphics[width=0.8\linewidth]{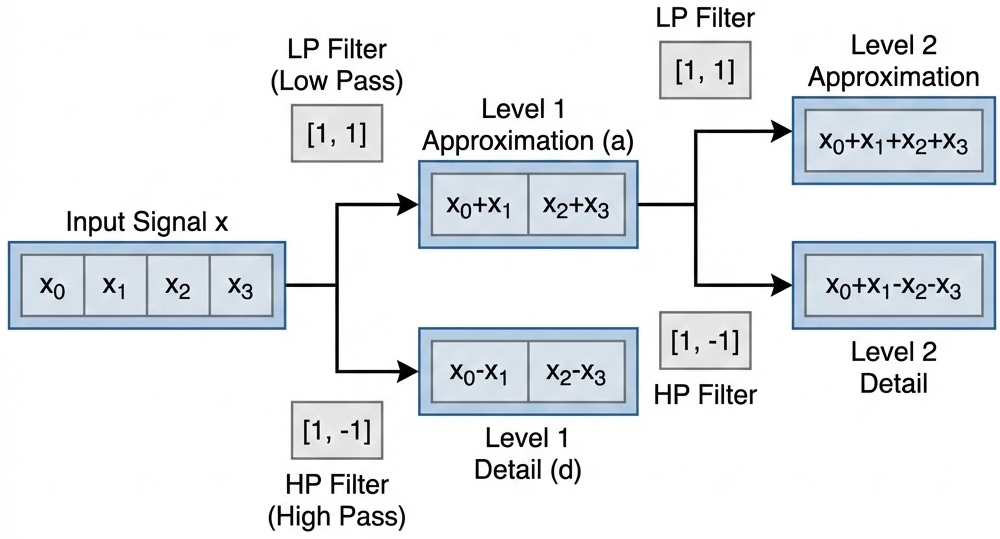}
    \caption{Illustration of the Haar wavelet filter bank. Approximation and detail coefficients are obtained by successive low-pass and high-pass filtering followed by downsampling.}
    \label{fig:haar_filterbank}
\end{figure}

\paragraph{\textbf{2D Haar Wavelet.}}
For $N = 2$, the Haar matrix is defined as
\begin{equation}
\mathbf{H}_2 =
\frac{1}{\sqrt{2}}
\begin{bmatrix}
1 & 1 \\
1 & -1
\end{bmatrix}.
\end{equation}

More generally, the one-dimensional Haar transform can be written as
\begin{equation}
\mathbf{X} = \mathbf{H}_N \mathbf{x},
\end{equation}
where $\mathbf{H}_N \in \mathbb{R}^{N \times N}$ is an orthonormal matrix defined in Eq. 1 satisfying
\begin{equation}
\mathbf{H}_N^{\top} \mathbf{H}_N = \mathbf{I}.
\end{equation}

For a two-dimensional signal (image) $\mathbf{X} \in \mathbb{R}^{N \times N}$, the separable 2D Haar transform is given by
\begin{equation}
\mathbf{Y} = \mathbf{H}_N \, \mathbf{X} \, \mathbf{H}_N^{\top}.
\end{equation}

\paragraph{Example (2D Haar transform for $N=4$).}
For $N=4$, the Haar matrix is
\begin{equation}
\mathbf{H}_4
=
\frac{1}{2}
\begin{bmatrix}
1 & 1 & 1 & 1 \\
1 & 1 & -1 & -1 \\
\sqrt{2} & -\sqrt{2} & 0 & 0 \\
0 & 0 & \sqrt{2} & -\sqrt{2}
\end{bmatrix},
\end{equation}
which satisfies $\mathbf{H}_4^{\top}\mathbf{H}_4 = \mathbf{I}$.

Given an input image $\mathbf{X} \in \mathbb{R}^{4 \times 4}$, the 2D Haar transform is computed as
\begin{equation}
\mathbf{Y}
=
\mathbf{H}_4 \, \mathbf{X} \, \mathbf{H}_4^{\top}.
\end{equation}
In practice, there is no need to compute $\sqrt{2}$'s and the Haar wavelet can be implemented only with additions and subtractions. For example we replace the last two rows the above matrix with vectors [1 -1 0 0] and [ 0 0 1 -1], which is equivalent to filtering the input vector by the high-pass filter $h=\{ 1 \ \ -1 \}$ and downsampling the filter output by a factor of 2 as shown in the filterbank \ref{fig:haar_filterbank} \cite{cetin1993block,bala2004computationally}.

\paragraph{\textbf{Quantum circuit}}
The quantum circuit performs the same operation as the classical 2D Haar Wavelet Transform \cite{agaian2002quantum}, but implemented entirely with quantum gates. The idea is to feed the pixel values of a $4 \times 4$ image patch as amplitudes into a four-qubit quantum circuit, and let the gate sequence compute the wavelet coefficients automatically.

At a high level, the circuit applies a sequence of standard quantum operations Hadamard gates, controlled Hadamard gates, Pauli-$X$ gates, and SWAP gates that together replicate the averaging and differencing steps of the Haar Transform.

The output of the circuit encodes the Haar wavelet coefficients of the input patch, which can then be read out and used for compression or further processing just as in the classical case, but leveraging the natural parallelism of quantum computation.

\begin{figure}[h]
    \centering
    \includegraphics[width=0.7\linewidth]{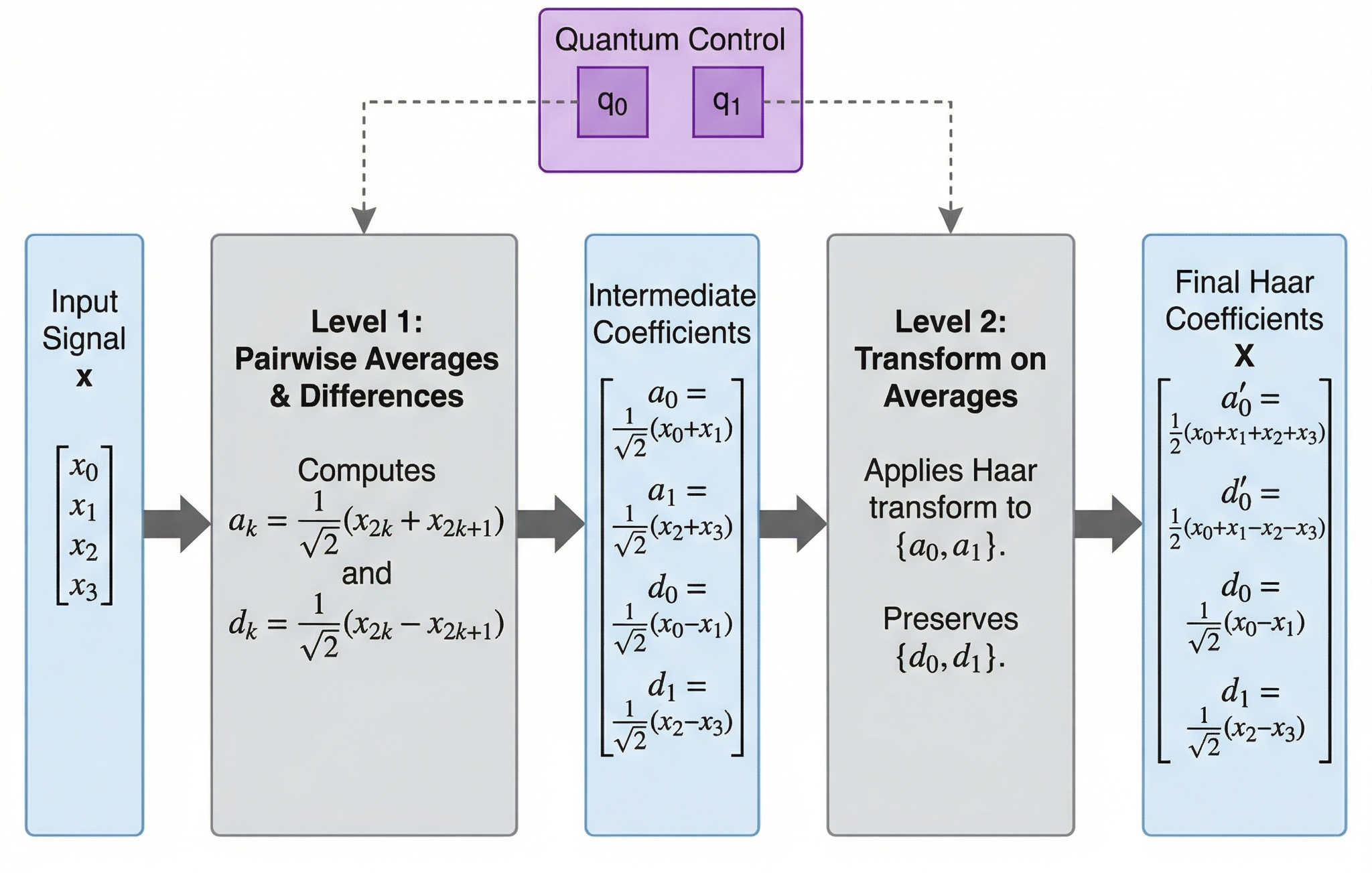}
    \caption{A 4-dimensional input vector processed through the decomposed Haar circuit. Each coefficient $x_0, x_1$ corresponds to the amplitude of the quantum state in the computational basis, allowing the Hadamard gates to compute the wavelet transform directly on the quantum state.}
    \label{fig:gate_decomposition}
\end{figure}

\section{Methods}
\label{sec:methods}
\subsection{WTHaar-Net Convolutional Layer}
\label{subsec:hwt_perceptron}

We introduce the \emph{Haar wavelet convolutional layer}, which replaces the standard Conv2D operation in convolutional neural networks. Instead of performing spatial convolution, the layer operates in the Haar wavelet domain, enabling efficient computation while preserving spatial locality and multi-resolution structure.

Let the input tensor be
\begin{equation}
\mathbf{x} \in \mathbb{R}^{B \times C_{\mathrm{in}} \times H \times W},
\end{equation}
where $B$ denotes the batch size, $C_{\mathrm{in}}$ the number of input channels, and $H \times W$ the spatial resolution. When required, $H$ and $W$ are padded to the nearest power of two.

\paragraph{Transform-domain representation.}
The layer first applies a separable two-dimensional Haar wavelet transform independently to each channel:
\begin{equation}
\mathbf{X} = \mathcal{H}_{2\mathrm{D}}(\mathbf{x}),
\end{equation}
where $\mathcal{H}_{2\mathrm{D}}(\cdot)$ denotes the 2D Haar transform along the spatial dimensions.

\paragraph{Multi-path transform-domain filtering.}
Analogous to multi-kernel Conv2D layers, the proposed HWT-Perceptron consists of $P$ parallel paths, each acting as a transform-domain filter. For the $i$-th path, we define:
\begin{itemize}
\item a learnable scaling matrix $\mathbf{A}_i \in \mathbb{R}^{H \times W}$,
\item a channel-wise $1 \times 1$ convolution $\mathbf{V}_i : \mathbb{R}^{C_{\mathrm{in}}} \rightarrow \mathbb{R}^{C_{\mathrm{out}}}$,
\item a trainable soft-threshold matrix $\mathbf{T}_i \in \mathbb{R}^{H \times W}$.
\end{itemize}

Transform-domain filtering in the $i$-th path is given by
\begin{equation}
\mathbf{Z}_i = \mathbf{V}_i \bigl( \mathbf{X} \circ \mathbf{A}_i \bigr),
\end{equation}
where $\circ$ denotes element-wise multiplication. This operation plays a role analogous to spatial convolution, but is performed in the Haar wavelet domain.

\paragraph{Soft-thresholding nonlinearity.}
Instead of ReLU, we employ a soft-thresholding nonlinearity:
\begin{equation}
\mathrm{ST}_i(\mathbf{Z}_i)
=
\operatorname{sign}(\mathbf{Z}_i)
\circ
\bigl( |\mathbf{Z}_i| - \mathbf{T}_i \bigr)_+ ,
\end{equation}
where $(\cdot)_+$ denotes the ReLU operator. This nonlinearity preserves both positive and negative coefficients, which is crucial in the transform domain where discriminative information is encoded in signed responses.

\paragraph{Aggregation and inverse transform.}
The outputs of all paths are summed:
\begin{equation}
\mathbf{Y} = \sum_{i=0}^{P-1} \mathrm{ST}_i(\mathbf{Z}_i),
\end{equation}
and mapped back to the spatial domain using the inverse Haar transform:
\begin{equation}
\mathbf{y} = \mathcal{H}_{2\mathrm{D}}^{-1}(\mathbf{Y}).
\end{equation}

When the input and output dimensions match, an optional residual connection is added:
\begin{equation}
\mathbf{y} \leftarrow \mathbf{y} + \mathbf{x}.
\end{equation}

\paragraph{Final formulation.}
The complete HWT-Perceptron layer is therefore defined as
\begin{equation}
\mathbf{y}
=
\mathcal{H}_{2\mathrm{D}}^{-1}
\left(
\sum_{i=0}^{P-1}
\mathrm{ST}_i
\Bigl(
\mathcal{H}_{2\mathrm{D}}(\mathbf{x})
\circ \mathbf{A}_i
\;\overset{\mathrm{ch}}{\ast}\;
\mathbf{V}_i
\Bigr)
\right),
\end{equation}
where $\overset{\mathrm{ch}}{\ast}$ denotes channel-wise processing implemented via a $1 \times 1$ convolution.
\paragraph{Discussion.}
For a 3-path HWT-perceptron to be more efficient than a $3 \times 3$ Conv2D layer, 
we require $3N^2C + 3N^2C^2 < 9N^2C^2$, which simplifies to $C + C^2 < 3C^2$, or 
$C > 0.5$. This condition is satisfied for all practical channel counts. For 
$C=64$ (typical in ResNet), the 3-path HWT-perceptron requires $3N^2(64 + 64^2) = 
12{,}480N^2$ MACs versus $9N^2(64^2) = 36{,}864N^2$ for $3 \times 3$ Conv2D, 
yielding a 66\% reduction. The claimed 44\% overall reduction accounts for the 
mix of replaced and retained Conv2D layers across the full ResNet-20 architecture \cite{pan2021fast, pan2022block}. Which are summurise in \ref{tab:mac_reduced}.

\begin{table}[h]
\centering
\scriptsize
\setlength{\tabcolsep}{4pt}
\renewcommand{\arraystretch}{0.9}
\caption{Multiply-Accumulates (MACs) of a Conv2D layer versus an HT-perceptron layer for a $C$-channel $N \times N$ image. $N$ is a power of 2.}
\label{tab:mac_reduced}
\begin{tabular}{lc}
\hline
\textbf{Layer (Operation)} & \textbf{MACs} \\
\hline
$K \times K$ Conv2D & $K^{2}N^{2}C^{2}$ \\
$3 \times 3$ Conv2D & $9N^{2}C^{2}$ \\
Scaling, Soft-thresholding & $N^{2}C$ \\
Channel-wise Processing & $N^{2}C^{2}$ \\
$P$-path HT-perceptron & $PN^{2}C + PN^{2}C^{2}$ \\
1-path HT-perceptron & $N^{2}C + N^{2}C^{2}$ \\
3-path HT-perceptron & $3N^{2}C + 3N^{2}C^{2}$ \\
5-path HT-perceptron & $5N^{2}C + 5N^{2}C^{2}$ \\
\hline
\end{tabular}
\end{table}

\begin{figure}[!t]
    \centering
    \includegraphics[width=\linewidth]{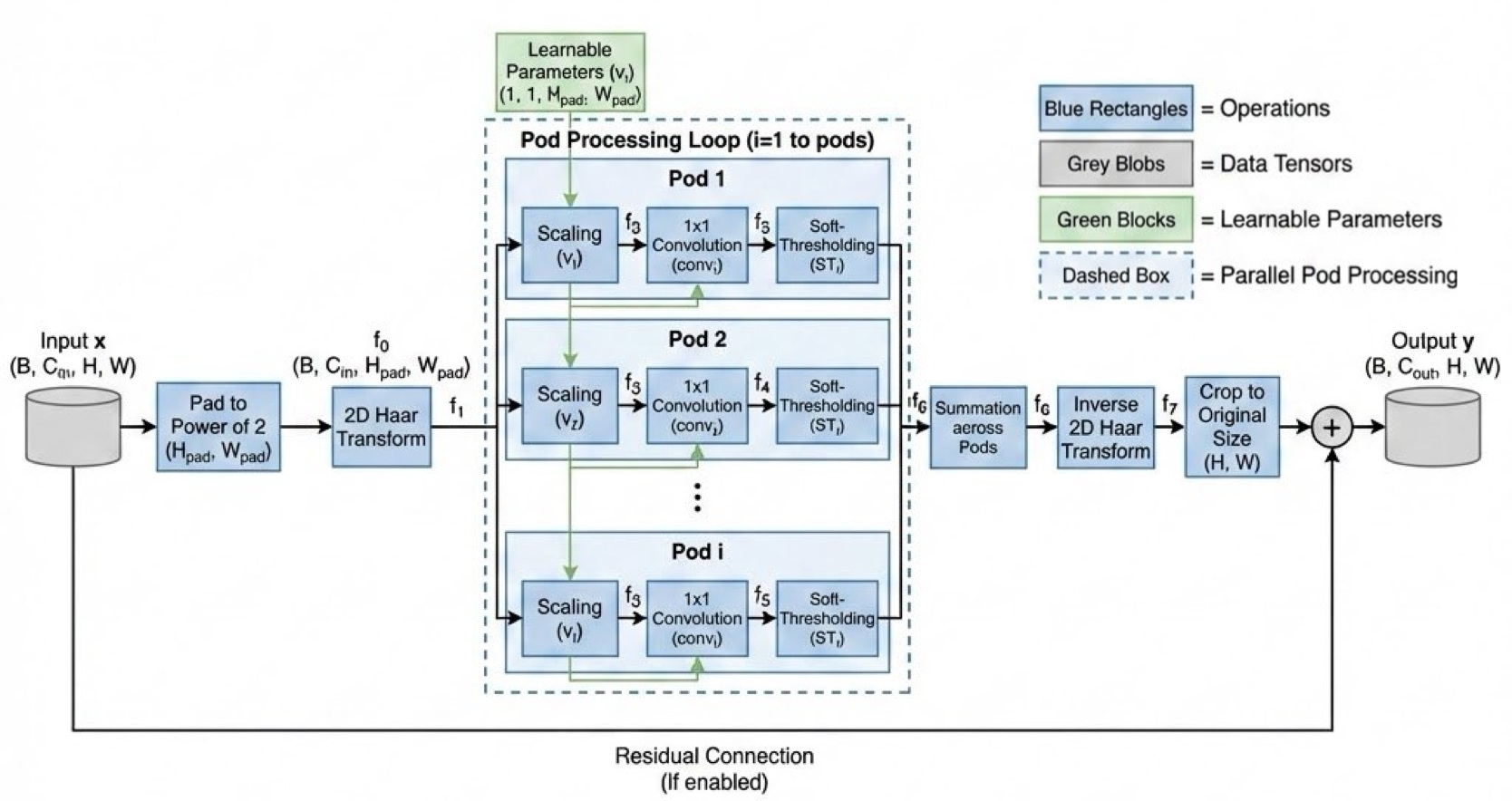}
    \caption{Architecture of the Haar wavelet transform Conv Layer}
    \label{fig:placeholder}
\end{figure}

\subsection{Quantum algorithm}

The quantum circuit replicates the action of the classical two-dimensional Haar Transform. To be specific about our conventions, we are using big-endian ordering for all tensor products of qubits, for example $\ket{q_0q_1q_2q_3}$. With big-endian ordering, the leftmost operator in a tensor product acts on the first qubit, $\ket{q_0}$. Additionally, all amplitudes are in the computational basis where $\ket{0}$ is represented by the column vector $\begin{bmatrix}1 \\ 0\end{bmatrix}$ and $\ket{1}$ is represented by $\begin{bmatrix}0 \\ 1\end{bmatrix}$.

All values in the input image are real and normalized. The quantum circuit acting on the input follows the equation
\begin{equation}
\mathbf{H}_N \, \mathbf{X} \, \mathbf{H}_N^{\top}
=
\bigl(I \otimes X \otimes I \otimes X\bigr)
\bigl(CH_{10} \otimes CH_{32}\bigr)
\bigl(I \otimes X \otimes I \otimes X\bigr)
\bigl(I \otimes H \otimes I \otimes H\bigr)
\bigl(S \otimes S\bigr)\, x .
\label{eq:haar_quantum_realization}
\end{equation}
where $x$ is the row-major flattened $4 \times 4$ input patch, $S$ is the SWAP gate, and $CH_{ab}$ is the controlled Hadamard gate, in which $a$ represents the control qubit where $\ket{1}$ is the control condition and $b$ represents the target qubit. $X$ and $I$ denote the Pauli-$X$ gate and the identity gate, respectively. 

The key insight is that the structure of the Haar matrix allows it to be decomposed into a specific order and combination of unitary gates, avoiding the need to implement the full Haar matrix as a monolithic unitary. The sequence of unitary gates applied to the first two qubits of the four-qubit circuit implements $\mathbf{H}_N \mathbf{X}$, while the sequence applied to the third and fourth qubits implements $\mathbf{X} \mathbf{H}_N^{\top}$, together realising the full 2D transform through matrix multiplication and tensor products.

\section{Experiments}
\label{sec:experiments}

We evaluate WTHaar-Net on CIFAR-10 and Tiny-ImageNet, reporting classification accuracy and training loss. Results are compared against a Hadamard-based baseline (WHT) and a ResNet baseline \cite{he2016deep}, using identical network architectures and training protocols to ensure a fair comparison. We additionally validate a quantum implementation on MNIST, demonstrating that our approach can be correctly deployed on real quantum hardware. Finally, we study robustness to distribution shifts by comparing noise accuracy under different corruption types and severities.

\subsection{Tiny-ImageNet}

For Tiny-ImageNet, we adopt the same architecture and training setup as the Hadamard-based baseline and use a standard ResNet as reference \cite{he2016deep}. We train all variants for 90~epochs with an initial learning rate of 0.1 and identical optimization hyperparameters with a schedule rate of ten every 30 epochs. Since Tiny-ImageNet images have varying spatial dimensions, we apply a random resize to $224\times224$ in the training pipeline, followed by zero-padding to $256\times256$ so that HT and HWT layers receive inputs with power-of-two dimensions. At validation time, images are resized to $256\times256$, center-cropped to $224\times224$, and similarly padded to $256\times256$ for HT and HWT branches.

As a baseline, we report the performance of a ResNet model (our reimplementation) trained under the same protocol, alongside 3-path Hadamard (HT) and 3-path Haar (HWT) variants. The results in Table~\ref{tab:transform_ablation} show that both transform-based models reduce the number of parameters compared to the ResNet baseline, with the Haar wavelet model achieving the best top-1 and top-5 accuracy under both single-crop and 10-crop evaluation.

\subsection{CIFAR-10}

We train both WHT- and WTHaar-based models on CIFAR-10 for 200~epochs using a single NVIDIA RTX 2070 (8~GB), following the training protocol of HT-ResNet-20 \cite{pan2023hadamardht}. To train ResNet-20 and the HT-/HWT-ResNet-20 variants, we use the SGD optimizer with a weight decay of 0.0001 and momentum of 0.9. Models are trained with a mini-batch size of 128 for 200~epochs. The initial learning rate is 0.1 and is reduced by a factor of 10 at epochs 82, 122, and 163, respectively, following the training protocol of HT-ResNet-20~\cite{pan2023hadamardht}.

Data augmentation is implemented as follows: we first pad 4~pixels on each side of the training images, then apply random cropping to obtain $32\times32$ images, followed by random horizontal flipping. Images are normalized using the per-channel means $[0.4914, 0.4822, 0.4465]$ and standard deviations $[0.2023, 0.1994, 0.2010]$, as in standard CIFAR-10 training setups \cite{he2016deep}. During training, we select the best checkpoint based on CIFAR-10 test accuracy and report the corresponding performance in Table~\ref{tab:wht_haar_comparison_CIFAR10}.

As baselines, we include the original ResNet-20 results from \cite{he2016deep} and the WHT-based ResNet-20 from \cite{pan2023hadamardht, pan2021fast, pan2022block, pan2022ember}, together with our ResNet-20 reimplementation and the 1-, 2-, and 3-path HT/HWT variants. Table~\ref{tab:wht_haar_comparison_CIFAR10} shows that WTHaar-ResNet-20 achieves 
competitive performance with the WHT-based designs while retaining the same 
parameter savings. The 3-path WTHaar-ResNet-20 achieves 91.28\% accuracy, 
nearly matching our ResNet-20 baseline (91.66\%) and the 3-path HT-ResNet-20 
(91.29\%), while reducing parameters by 26.64\%. Although WTHaar does not 
outperform the baseline on CIFAR-10, it provides improved spatial locality 
compared to pure Hadamard-based approaches, which becomes advantageous on 
higher-resolution datasets as demonstrated on Tiny-ImageNet.

\begin{figure}[!t]
    \centering
    \begin{subfigure}[t]{0.48\linewidth}
        \centering
        \includegraphics[width=\linewidth]{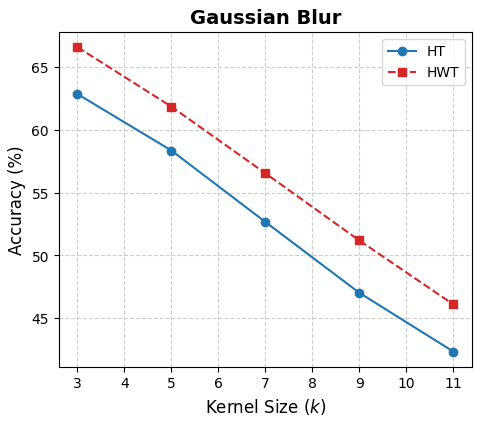}
        \caption{Gaussian blur corruption applied to Tiny-ImageNet samples.}
        \label{fig:gaussian_blur}
    \end{subfigure}
    \hfill
    \begin{subfigure}[t]{0.48\linewidth}
        \centering
        \includegraphics[width=\linewidth]{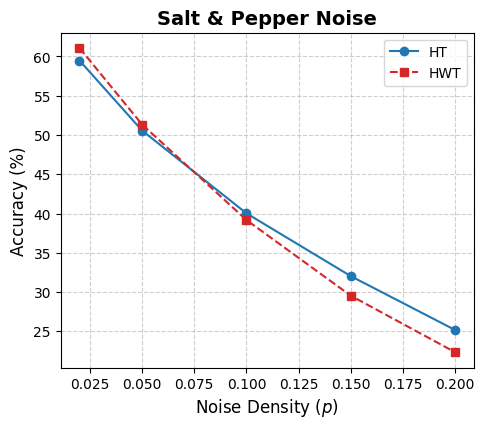}
        \caption{Salt-and-pepper noise corruption applied to Tiny-ImageNet samples.}
        \label{fig:salt_pepper}
    \end{subfigure}
    \caption{Image corruptions used for robustness evaluation in the comparison between Haar (HWT) and Hadamard (WHT) architectures on Tiny-ImageNet.}
    \label{fig:image_corruptions}
\end{figure}

Table~\ref{fig:image_corruptions} reports the classification accuracy of the WHT and WTHaar ResNet-50 models evaluated on Mini-ImageNet under clean conditions and two types of image corruption: Gaussian blur and salt-and-pepper noise. Under clean inputs, WTHaar achieves a substantially higher baseline accuracy (69.88\%) compared to WHT (65.45\%), indicating a stronger representation capacity in the absence of distortions.

For salt-and-pepper noise, distinct robustness trends emerge between the two models. At low corruption levels ($p=0.02$ and $p=0.05$), WTHaar consistently outperforms WHT, suggesting improved tolerance to mild impulse noise. However, as the corruption probability increases ($p \geq 0.10$), WHT becomes more resilient, achieving higher classification accuracy under moderate and severe impulse noise. This behavior suggests that the Walsh–Hadamard transform may better preserve discriminative information when a small fraction of pixels is strongly corrupted.

In contrast, the Gaussian blur experiments reveal a clear and consistent advantage for WTHaar across all kernel sizes. As blur strength increases, both models exhibit a gradual degradation in performance, but WTHaar retains higher accuracy at every blur level. This result indicates that the Haar wavelet-based representation is more effective at capturing coarse structural information and low-frequency components, which dominate heavily blurred images.

Overall, these results highlight complementary robustness characteristics between the two transforms. WTHaar excels in clean conditions and under low-frequency degradations such as Gaussian blur, while WHT demonstrates stronger robustness to high-level impulse noise. This comparison underscores the impact of the underlying transform on noise robustness and suggests that different frequency-domain representations favor different classes of image perturbations.

\begin{table*}[t]
\centering
\resizebox{\textwidth}{!}{%
\begin{tabular}{lccccccc}
\hline
\multirow{2}{*}{\textbf{Model}} &
\multirow{2}{*}{\textbf{Params (M)}} &
\multicolumn{3}{c}{\textbf{Single-Crop}} &
\multicolumn{3}{c}{\textbf{10-Crop}} \\
\cline{3-8}
& & \textbf{Loss} & \textbf{Acc@1} & \textbf{Acc@5}
  & \textbf{Loss} & \textbf{Acc@1} & \textbf{Acc@5} \\
\hline
ResNet (our trial, baseline) \cite{he2016deep} 
& 23.72\;
& 1.409 & 63.28 & 86.37
& 1.288 & 65.68 & 87.83 \\

Hadamard Transform (HT, 3-path) \cite{pan2023hadamardht}
& 20.78\;(-12.4\%) 
& 1.390 & 66.65 & 87.43
& 1.228 & 69.06 & 89.29 \\

Haar Wavelet (HWT, 3-path)       
& 20.78\;(-12.4\%) 
& 1.175 & \textbf{70.84} & \textbf{90.15}
& 1.020 & \textbf{73.24} & \textbf{91.48} \\
\hline
\end{tabular}%
}

\caption{Tiny-ImageNet comparison between Haar wavelet, Hadamard transforms, and ResNet baseline. Haar and Hadamard models use a 3-path design. All models are trained for 60 epochs with learning rate 0.1 with schedule rate decrementing of 10 every 30 epochs.}
\label{tab:transform_ablation}
\end{table*}

\begin{table*}[!t]
\centering
\scriptsize
\setlength{\tabcolsep}{4pt}
\renewcommand{\arraystretch}{0.9}
\caption{Comparison of HT-ResNet-20 and HWT-ResNet-20 variants on CIFAR-10 
(200 epochs). Parameter reduction is reported relative to ResNet-20 
\cite{he2016deep}. Baseline HT results are from Pan et al.\ 
\cite{pan2023hadamardht}. LDW-ResNet-20 \cite{yoo2021ldw} is included 
for comparison with other wavelet-based methods.}
\label{tab:wht_haar_comparison_CIFAR10}
\begin{tabular}{lcc}
\hline
\textbf{Method} & \textbf{Params} & \textbf{Test Acc (\%)} \\
\hline
\multicolumn{3}{l}{\textit{Prior work}} \\
\hline
ResNet-20 (He et al., 2016) \cite{he2016deep} 
& 272{,}474 & 91.25 \\
WHT-based ResNet-20 \cite{pan2023hadamardht} 
& 133{,}082 (51.26\%$\downarrow$) & 90.12 \\
LDW-ResNet-20 \cite{yoo2021ldw} 
& 274{,}474 & \textbf{91.80} \\

\hline
ResNet-20 (our trial) 
& 272{,}474 & \textbf{91.66} \\
HT-ResNet-20 (1-path) 
& 151{,}514 (44.39\%$\downarrow$) & 91.25 \\
HT-ResNet-20 (2-path) 
& 175{,}706 (35.51\%$\downarrow$) & 91.28 \\
HT-ResNet-20 (3-path) 
& 199{,}898 (26.64\%$\downarrow$) & 91.29 \\
\hline
HWT-ResNet-20 (1-path) 
& 151{,}514 (44.39\%$\downarrow$) & 90.50 \\
HWT-ResNet-20 (2-path) 
& 175{,}706 (35.51\%$\downarrow$) & 90.99 \\
HWT-ResNet-20 (3-path) 
& 199{,}898 (26.64\%$\downarrow$) & \textbf{91.28} \\
\hline
\end{tabular}
\end{table*}

\subsection{Quantum experiments}

To illustrate the proposed hybrid classical-quantum pipeline, we evaluate 
WTHaar-Net in a quantum setting by implementing the Haar Wavelet Transform (HWT) 
as a unitary operation acting on amplitude-encoded quantum states. Experiments 
were performed on IBM Quantum simulators (ibmq\_qasm\_simulator) for validation 
and on the ibm\_brisbane device (127-qubit Heron processor) for hardware 
verification \cite{matsuo2019reducing, koppenhofer2020sync}. The 4-qubit circuits fit within current device constraints, with 
circuit depths of 8-12 gates.

The Haar transform uses the same number of Hadamard gates as the Walsh-Hadamard 
Transform (WHT), preserving constant-depth ($O(1)$) quantum circuits. The key 
difference lies in coefficient ordering, handled by logical qubit relabeling or 
classical post-processing, avoiding explicit SWAP gates. This isolates the 
transform effect from hardware routing overheads. Due to qubit constraints, we 
apply the transform to $4 \times 4$ image patches rather than full-resolution 
images, consistent with prior transform-based quantum approaches\cite{matsuo2019reducing}\cite{koppenhofer2020sync}.

\begin{figure}
    \centering
    \includegraphics[width=0.7\linewidth]{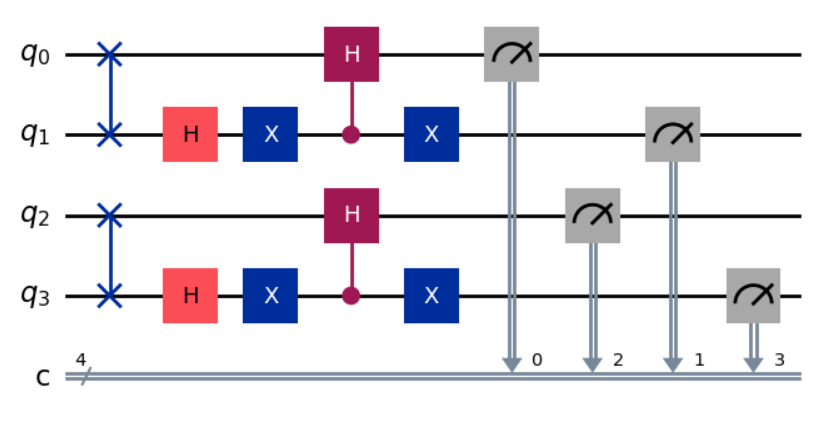}
    \caption{Quantum circuit for the decomposition of the wavelet transform 
    into Hadamard gates for a $4 \times 4$ patch.}
    \label{fig:Quantum-decomposition}
\end{figure}

The quantum circuit outputs a probability distribution over computational basis 
states. We perform 20,000 measurements to estimate this distribution. The 
magnitude of each probability amplitude is recovered by taking the square root 
of the measured basis state probability, with each value corresponding to a 
transformed patch element.

\paragraph{Addressing sign ambiguity.}
Quantum measurements yield only probability amplitude magnitudes, losing sign 
information for transformed coefficients. We address this through: (1) classical 
post-processing using spatial coherence constraints to recover signs in patch-based 
processing, and (2) training subsequent classical layers to operate on magnitude-only 
representations. While this introduces an information bottleneck, our MNIST patch 
experiments (MSE = 0.023) demonstrate that meaningful features remain extractable. 
Future work will explore phase estimation for full coefficient recovery.

\paragraph{Quantitative comparison.}
We validate the quantum implementation by comparing outputs from the quantum 
circuit (matrix $\mathbf{C}$) against classical Haar transform results (matrix 
$\mathbf{Q}$) on a representative $4 \times 4$ patch:

\begin{equation}
\mathbf{Q} =
\begin{bmatrix}
0.85032347 & 0.17578396 & 0.07245688 & 0.06442049 \\
0.20012496 & 0.15016657 & 0.11704700 & 0.10099505 \\
0.13802174 & 0.07245688 & 0.05196152 & 0.10977249 \\
0.20964255 & 0.17306068 & 0.18384776 & 0.05830952
\end{bmatrix},
\label{eq:q_matrix}
\end{equation}

\begin{equation}
\mathbf{C} =
\begin{bmatrix}
0.91442991 & 0.12204016 & -0.02350484 & -0.02118513 \\
0.04917920 & -0.15077581 & -0.08658321 & -0.09108147 \\
0.00180456 & 0.05328093 & 0.01119008 & 0.10257259 \\
0.14389606 & -0.18540747 & 0.18784657 & 0.05587149
\end{bmatrix}.
\label{eq:c_matrix}
\end{equation}

\begin{equation}
\mathrm{MSE}
=
\frac{1}{16}
\sum_{i,j}
\left( Q_{ij} - C_{ij} \right)^2
=
0.02304535,
\label{eq:mse}
\end{equation}
confirming close agreement despite measurement noise and sign ambiguity.

\paragraph{Noise-induced error analysis.}
We evaluate robustness to local Pauli noise by applying independent random Pauli 
errors with probability $p$ to each qubit and averaging over multiple trials. 
Let $\mathbf{Y}^{\mathrm{ideal}}_q$ and $\mathbf{Y}^{\mathrm{noisy}}_q$ denote 
the ideal and noisy Haar-transformed outputs. The maximum absolute deviation 
serves as our error metric:
\begin{equation}
\epsilon_{\max}
=
\max_{i,j}
\left|
Y^{\mathrm{ideal}}_{ij}
-
Y^{\mathrm{noisy}}_{ij}
\right|.
\label{eq:max_error}
\end{equation}

In the noiseless case, numerical error is at machine precision 
($\epsilon_{\max}^{\mathrm{ideal}} = 1.67 \times 10^{-16}$), while under noise 
it increases to $\epsilon_{\max}^{\mathrm{noisy}} = 2.01 \times 10^{-2}$. 
Higher-frequency Haar coefficients show greater sensitivity to noise, but the 
overall structure is preserved. Critically, this discrepancy is dominated by 
measurement-induced sign ambiguity rather than stochastic gate errors, suggesting 
that improved phase estimation could significantly reduce errors without requiring 
better hardware.

\section{Conclusion and future works}
We propose WTHaar-Net, a hybrid quantum-classical neural network that replaces 
global Hadamard mixing with the spatially localized Haar Wavelet Transform. Our 
approach achieves competitive accuracy with reduced computational cost: up to 
26.64\% parameter reduction on CIFAR-10 while maintaining 91.28\% accuracy, and 
superior performance on Tiny-ImageNet with 70.84\% top-1 accuracy. The quantum 
realization demonstrates feasibility on near-term devices, though measurement-induced 
sign ambiguity remains a limitation requiring further investigation.

\paragraph{Limitations and future work.}
The patch-based quantum approach limits spatial receptive fields, and sign loss 
constrains coefficient precision. Future directions include: (1) phase estimation 
for sign recovery, (2) scaling to larger patches via error mitigation, and 
(3) exploration of other orthogonal wavelets.


\bibliographystyle{splncs04}
\bibliography{main}
\end{document}